\documentclass[11pt]{article}

\usepackage[preprint]{acl}
\usepackage{times}
\usepackage{latexsym}
\usepackage{booktabs}
\usepackage{amsmath}
\usepackage{amssymb}
\usepackage{enumitem}
\usepackage{subcaption}
\author{
\normalfont
Akshata A. Bhat\textsuperscript{1},
Daman Deep Singh\textsuperscript{1},
Saiful Haq\textsuperscript{1,2},
Niyati Chhaya\textsuperscript{1} \\[4pt]
\textsuperscript{1}Hyperbots Inc. \\
\textsuperscript{2}Indian Institute of Technology Bombay
}
\usepackage[T1]{fontenc}

\usepackage[utf8]{inputenc}

\usepackage{microtype}

\usepackage{inconsolata}

\usepackage{comment}
\usepackage{multirow}

\usepackage{graphicx}
\usepackage{tikz}
\usetikzlibrary{shapes,arrows,positioning}
\usepackage{graphicx}
\usepackage{booktabs}
\usepackage{tikz}
\usepackage{xcolor}
\usetikzlibrary{arrows.meta,positioning,calc,fit,backgrounds,shapes.geometric,decorations.pathreplacing}
\usepackage{pifont}
\definecolor{replcolor}{RGB}{0,120,180}
\definecolor{delcolor}{RGB}{200,50,50}
\definecolor{inscolor}{RGB}{40,160,60}
\definecolor{docbg}{RGB}{255,252,245}
\definecolor{tableborder}{RGB}{160,160,160}

\newcommand{\docintent}{\texttt{DocIntent}}
\newcommand{\intentbench}{\texttt{IntentBench}}
\newcommand{\docintentocr}{\texttt{DocIntentOCR}}

\title{When Ink Meets Pixel: Intent-Aware OCR with VLMs for Human-Corrected Financial Documents}

\begin{document}
\maketitle            
\begin{abstract}
Human annotations on documents--such as arrows, carets, strikethroughs, and insertions--express corrective intent yet existing OCR and document understanding systems treat these marks as visual noise rather than semantic edit operations. We introduce \textit{intent-aware OCR}, the task of generating corrected transcriptions by jointly reasoning over original document content and visual intent cues (handwritten or digital) that convey edit intent. To support this setting, we present \texttt{DocIntent}, a multimodal dataset of 1,600 financial documents with over 5,700 annotated visual intent cues, each paired with layout-aware, intent-resolved OCR targets. We further introduce \texttt{IntentBench}, a 216-document benchmark designed to evaluate model robustness under diverse real-world annotation conditions. Leveraging \texttt{DocIntent}, we fine-tune \texttt{DocIntentOCR}, a suite of 3B and 7B parameter vision-language models, under direct intent supervision on corrected transcriptions. On \texttt{IntentBench}, the fine-tuned 3B and 7B models achieve F1-scores of 0.81 and 0.83, respectively -- substantially outperforming commercial closed-source (e.g., GPT-4.1 (0.626)), and open-source models (e.g., LLaMA-3.2-Vision-90B (0.50)) demonstrating that intent-aware supervision enables compact domain-specific models to outperform substantially larger general-purpose systems on this task. This work establishes intent-aware OCR as a novel human-in-the-loop document understanding task and provides foundational resources (dataset, benchmark, models) to advance research in interactive document processing.

\end{abstract}
  
\section{Introduction}

When an accountant draws a strikethrough across an incorrect line item on an invoice, or a compliance officer inserts a corrected figure with a caret symbol, these marks constitute a decision -- a deliberate instruction to modify the document. Across finance, legal, and auditing workflows, printed and digital documents routinely pass through human hands, accumulating annotations that encode corrective intent: deletions, replacements, insertions, and verifications. These visual cues -- strikethroughs, arrows, carets, and circled values -- form a rich, informal language through which reviewers communicate how document content should be changed. The scale of this problem is significant: of the estimated 550 billion invoices produced globally each year~\cite{koch2019einvoicing}, fewer than one-third achieve straight-through processing without human intervention~\cite{ardent2024epayables}, leaving the majority subject to manual review workflows where such annotations accumulate.

Optical Character Recognition (OCR) systems, from traditional engines such as Tesseract~\cite{smith2007tesseract} to modern Vision-Language Models (VLMs) like LayoutLM~\cite{xu2020layoutlm}, Donut~\cite{kim2022donut}, DeepSeek-OCR~\cite{wei2025deepseekocr} are designed to extract text while being invariant to visual noise. 
% This invariance, a strength for clean documents, becomes a critical weakness for annotated ones: the very marks that encode human corrections are treated as distortions to be suppressed. The result is a fundamental disconnect: OCR output reflects the document as it was in its original, unannotated state, not as the human \emph{intends} it to be, due to treatment of visual cues as noise. In practice, this forces organizations to rely on additional manual correction or brittle rule-based post-processing pipelines that operate outside the visual document space~\cite{barrachina2009imt}.
The result is a fundamental disconnect: OCR output reflects the original, unannotated document, not the human \emph{intended} version, because visual cues are treated as noise. In practice, this forces organizations to rely on additional manual correction or brittle rule-based post-processing pipelines that operate outside the visual document space~\cite{barrachina2009imt}.

% We argue that this limitation reflects a missing abstraction: existing OCR formulations have no mechanism for modeling \textit{human intent}. Human annotations (handwritten or digital) are not arbitrary marks--they are visual edit commands. A strikethrough means \emph{delete this}. A caret with text means \emph{insert here}. An arrow means \emph{move this there}. We formalize this insight as \emph{intent-aware OCR}: a document analysis task in which transcription is conditioned on both printed content and visual intent cues--whether handwritten or digital--interpreted as semantic edit operations. Depending on annotation scope, these operations may span tokens, fields, table cells, rows, or entire columns.

We argue that this limitation reflects a missing abstraction: existing OCR formulations lack a mechanism for modeling \textit{human intent}. Annotations (handwritten or digital) are not arbitrary marks—they are visual edit commands: strikethroughs indicate \emph{deletions}, carets indicate \emph{insertions}, and arrows indicate \emph{moves}. We formalize this as \emph{intent-aware OCR}, a document analysis task in which transcription is conditioned on both printed content and visual intent cues, interpreted as semantic edit operations. Depending on scope, these operations may span tokens, fields, table cells, rows, or entire columns.

% To enable systematic study of this task, we introduce \docintent{}, a multimodal dataset of 1,600 synthetic financial documents augmented with over 5,700 visual intent cues, each paired with a layout-aware, intent-resolved OCR target. For realistic evaluation, we present \intentbench{}, a 250-document benchmark of real-world annotated financial documents. Leveraging these resources, we fine-tune \docintentocr{} -- a family of 3B and 7B models -- under direct intent supervision and show that these compact domain-specific models substantially outperform significantlly larger closed-source and open-source models.

To enable systematic study of this task, we introduce \docintent{}, a multimodal dataset of 1,600 synthetic financial documents augmented with over 5,700 visual intent cues, each paired with a layout-aware, intent-resolved OCR target. For realistic evaluation, we present \intentbench{}, a 216-document benchmark of real-world annotated financial documents. Leveraging these resources, we fine-tune \docintentocr{}—a family of 3B and 7B models—under direct intent supervision, showing that these compact, domain-specific models achieve up to 0.83 F1, substantially outperforming significantly larger closed- and open-source models, including GPT-4.1 (0.626) and LLaMA-3.2-Vision-90B (0.50).

In summary, this work makes the following contributions:
\begin{itemize}[noitemsep]
    \item We formalize \emph{intent-aware OCR}, a task that generates corrected transcriptions by interpreting visual feedback cues (e.g., arrows, strikethroughs, insertions) as edit operations.
    \item We introduce \docintent{} (1,600 documents, 5,700+ intent cues) and \intentbench{} (216 real-world documents), the first dataset and benchmark for intent-aware OCR.
    \item We fine-tune \docintentocr{} under direct supervision, showing that intent-aware fine-tuning enables compact VLMs to outperform much larger models on this task.
\end{itemize}

By treating handwritten annotations as executable edits rather than noise, this work opens a new direction in document understanding -- one where OCR systems become participants in, rather than bystanders to, human review workflows.

\section{Related Work}

% OCR research has progressed from rule-based engines~\cite{smith2007tesseract} through layout-aware pretrained models~\cite{xu2020layoutlm,xu2021layoutlmv2,huang2022layoutlmv3,hong2022bros} to end-to-end architectures~\cite{kim2022donut,blecher2023nougat,lee2022pix2struct} and unified models such as GOT-OCR~\cite{wei2024gotocr} and DeepSeek-OCR~\cite{wei2025deepseekocr}. General-purpose VLMs~\cite{bai2025qwen25vl,xiao2024florence2} and document-specialized architectures~\cite{liu2024textmonkey,hu2024docowl15} achieve strong results on benchmarks like DocVQA~\cite{mathew2021docvqa}. All these systems optimize for text fidelity---handwritten annotations are transcribed verbatim or suppressed as noise, never interpreted as edit operations that modify surrounding content. Offline handwriting recognition~\cite{marti2002iam,gervais2024mathwriting} and strikethrough-removal systems~\cite{chaudhuri2017strikethrough,heil2021strikethrough} similarly treat annotations as signal to recognize or discard.

OCR research has progressed from rule-based engines~\cite{smith2007tesseract} to layout-aware pretrained models~\cite{xu2020layoutlm,huang2022layoutlmv3}, end-to-end architectures~\cite{kim2022donut,blecher2023nougat}, and unified models such as GOT-OCR~\cite{wei2024gotocr} and DeepSeek-OCR~\cite{wei2025deepseekocr}. General-purpose VLMs~\cite{bai2025qwen25vl} and document-specialized architectures~\cite{liu2024textmonkey} achieve strong results on benchmarks like DocVQA~\cite{mathew2021docvqa}. These systems optimize for text fidelity—handwritten annotations are transcribed verbatim or suppressed as noise, never interpreted as edit operations. Offline handwriting recognition~\cite{marti2002iam,gervais2024mathwriting} and strikethrough-removal systems~\cite{heil2021strikethrough} similarly treat annotations as signal to recognize or remove.

% Human-in-the-loop paradigms such as active learning~\cite{settles2009activelearning} and interactive machine translation~\cite{barrachina2009imt} incorporate user feedback through iterative refinement cycles applied \emph{after} a model produces output. Intent-aware OCR inverts this relationship: corrective intent is already embedded in the input as visual marks on the document, and the model must recover it from a single static image without iterative clarification.
Human-in-the-loop paradigms such as active learning~\cite{settles2009activelearning} and interactive machine translation~\cite{barrachina2009imt} incorporate user feedback through iterative refinement \emph{after} a model produces output. Intent-aware OCR inverts this: corrective intent is embedded directly in the input as visual annotations, and the model must recover it from a single static image without iterative clarification.

\section{Intent-Aware OCR: Task Definition}

This section formalizes intent-aware OCR as a document analysis task, defines the space of visual intent cues and their induced edit operations, and specifies the evaluation protocol.

\subsection{Formulation}

Let $I$ denote a document image containing printed text, layout structure, and zero or more handwritten or digital annotations. A conventional OCR system maps $I$ to a transcription $T$ that faithfully reproduces all printed content. Intent-aware OCR instead maps $I$ to an \emph{intent-resolved transcription} $O$ that reflects modifications specified by visual annotations. Importantly, the formulation is agnostic to the serialization format of $O$: it applies equally whether the output is a layout-aware transcription that preserves two-dimensional document structure, a linear reading-order rendering, or any other text representation. The edit operations defined below operate over logical document regions (tokens, cells, rows, columns, tables) and are therefore independent of how the final transcription is linearized.

We model this as follows. Let $T$ denote the base transcription of the printed content, and let $\mathcal{H} = \{h_1, h_2, \ldots, h_n\}$ denote the set of visual intent cues present in the image. Each cue $h_i$ is associated with:
\begin{enumerate}[noitemsep]
    \item a \textbf{cue type} $c_i$ drawn from the taxonomy in Table~\ref{tab:intent-taxonomy} (Appendix~\ref{app:taxonomy}),
    \item a \textbf{structural scope} $s_i$ specifying the granularity of the edit (token, field, cell, row, column, or table), and
    \item an optional \textbf{replacement value} $v_i$ (for replacement and insertion operations).
\end{enumerate}

\noindent Each cue $h_i$ induces a deterministic edit operation $e_i = (c_i, s_i, v_i)$ over the base transcription. The intent-resolved output is defined as the sequential application of all edit operations:
\[
O = e_n \circ e_{n-1} \circ \cdots \circ e_1(T)
\]

In our formulation, each edit $e_i$ targets a spatially distinct region of the base transcription (a specific token span, cell, row, column, or table). When edit scopes are pairwise disjoint \footnote{This holds for all samples in our proposed datasets.}, the composition is commutative and the ordering is immaterial. In principle, overlapping scopes can arise (e.g., a row-level deletion and a cell-level replacement within the same row); we treat such cases as annotation ambiguities outside the scope of this taxonomy, and they do not occur in \docintent{} or \intentbench{}.

Critically, the model does not observe $T$ or $\mathcal{H}$ directly--it must infer both from the raw image $I$. The task therefore requires the model to simultaneously: (1)~recognize the printed document content, (2)~detect and classify visual annotations, (3)~determine the structural scope of each annotation, and (4)~apply the induced edits to produce the final output.

\paragraph{Handling Out-of-Taxonomy Annotations.} The taxonomy in Table~\ref{tab:intent-taxonomy} (Appendix~\ref{app:taxonomy}) does not exhaustively cover every handwritten annotation convention (e.g., circling, underlining, or marginal comments); these were deliberately excluded because they are structurally ambiguous--a circle, for instance, can indicate deletion, emphasis, or a query depending on context--and including them without an operationally unambiguous definition would undermine consistent supervision. For an annotation that does not match any of the ten defined cue types, the expected (and current) model behavior is a \emph{no-op}: the underlying text is retained as-is, and the model is expected to flag the unmatched region in its output rather than silently guessing an edit. We discuss the disambiguation challenges this introduces, and the resulting coverage limitation, in the Limitations section.

\subsection{Edit Operation Taxonomy}

We define three high-level categories of intent-induced transformations:
\begin{itemize}[noitemsep]
    \item \textbf{Deletion:} existing text or structured content is marked for removal (e.g., a strikethrough with no replacement value).
    \item \textbf{Replacement:} existing content is invalidated and substituted with a corrected value (e.g., a strikethrough accompanied by a handwritten correction).
    \item \textbf{Insertion:} new content is introduced at a specified position without removing existing text (e.g., a caret with handwritten text).
\end{itemize}

% Each category may operate at varying structural granularities -- from individual tokens to entire table rows, columns, or tables. A key property is that the same visual mark maps to different operations depending on context: a strikethrough alone induces deletion, whereas a strikethrough accompanied by a nearby handwritten value induces replacement. Table~\ref{tab:intent-taxonomy} (Appendix~\ref{app:taxonomy}) enumerates all cue types, structural scopes, and expected OCR behaviors.

Each category may operate at varying structural granularities—from individual tokens to entire rows, columns, or tables. A key property is that the same visual cue maps to different operations depending on context: a strikethrough alone induces deletion, whereas a strikethrough with a nearby handwritten value induces replacement. Table~\ref{tab:intent-taxonomy} (Appendix~\ref{app:taxonomy}) enumerates all cue types, scopes, and expected OCR behaviors, and Figure~\ref{fig:cue-gallery} (Appendix~\ref{app:taxonomy}) shows a real annotated example of each.

\subsection{Illustrative Example}

\begin{figure}[h]
\centering
\includegraphics[width=0.5\textwidth]{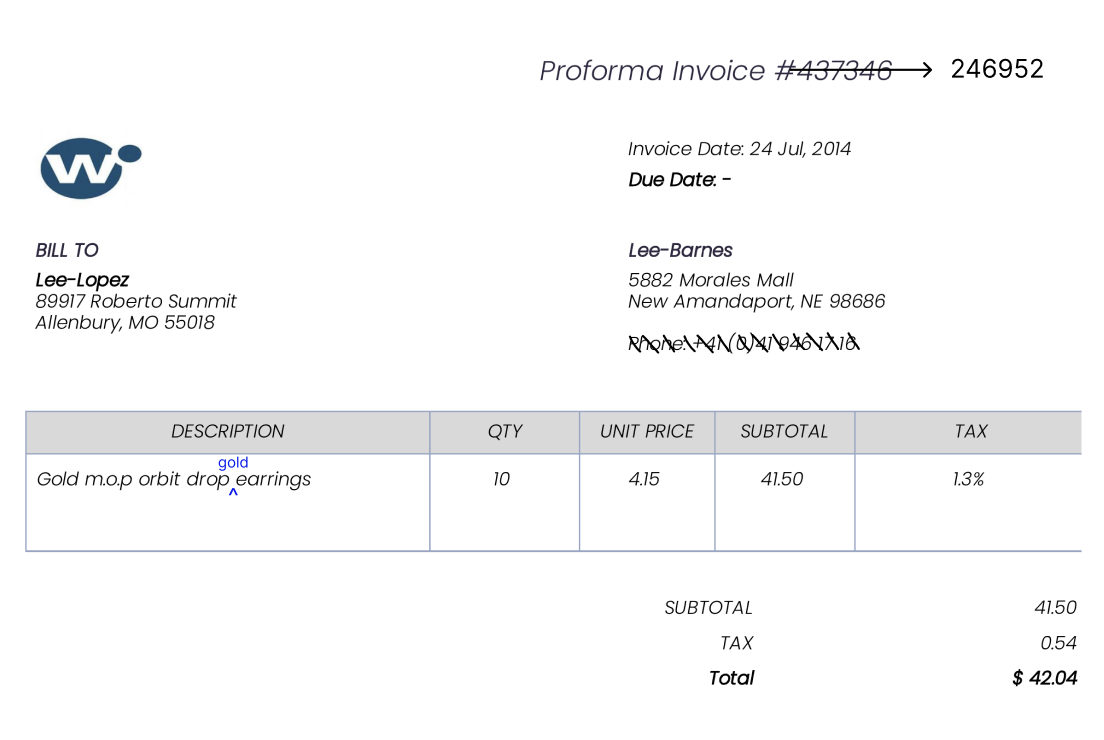}
\caption{A sample annotated invoice from \docintent{}. Three intent cues are present.}
\label{fig:task-overview}
\end{figure}

Figure~\ref{fig:task-overview} illustrates the task on a sample from \docintent{}. A human reviewer has: (1)~struck through the invoice number \texttt{\#437346} and drawn an arrow pointing to the handwritten replacement \texttt{246952} (arrow-indicated replacement at the token level), (2)~placed a caret below the word ``drop'' in the item description with the handwritten text ``gold'' (insertion at the token level), and (3)~scribbled out the phone number (deletion at the token/field level). Standard OCR transcribes the document verbatim and ignores these annotations. Intent-aware OCR interprets each annotation as an edit operation and produces a corrected transcription, outputting \texttt{246952} in place of the original invoice number, expanding the description to ``Gold m.o.p orbit drop \textit{gold} earrings,'' and omitting the phone number.

\subsection{Output Format}

The model produces a \emph{layout-aware plain-text transcription} that serializes document content according to a two-dimensional spatial grid, where each word is positioned by its row and column index, preserving table boundaries, row/column associations, and reading order~\cite{bhat2026savior}. We adopt this format because financial documents are inherently structured: a value in the wrong row or column constitutes a semantically incorrect transcription even when every individual token is recognized correctly. A flat, reading-order output, while human-readable, loses the spatial relationships that downstream key-value extraction and tabular parsing depend on.

\subsection{Evaluation Protocol}

We evaluate by comparing predicted intent-resolved transcriptions against ground-truth outputs using token-level F1-score, computed over the \emph{complete} intent-resolved output transcription against the complete ground truth--following established practice in document understanding benchmarks such as DocVQA~\cite{mathew2021docvqa}--rather than over an extracted span or the cue regions alone. This is a natural fit for a task whose intent errors propagate: a missed deletion or an incorrect replacement value alters the surrounding transcription, not just the annotated token, so all output tokens are scored uniformly regardless of whether they correspond to an intent cue or to unannotated content. Because token-level F1 by itself cannot isolate sequence-level degradation from cascading errors (e.g., a misclassified cue shifting subsequent tokens), we additionally report Word Error Rate (WER), computed as the standard Levenshtein-based word error rate over the linearized intent-resolved transcription (lower is better). F1 captures edit-operation correctness at the token level, while WER captures sequence-level transcription quality; together they give a more complete picture of intent-aware OCR performance than either alone. All evaluation is performed on \intentbench{}.

\section{Datasets}

We introduce two complementary resources: \docintent{}, a large-scale synthetic dataset for training, and \intentbench{}, a real-world benchmark for evaluation. Both are constructed around the visual intent cue taxonomy (Appendix~\ref{app:taxonomy}). Table~\ref{tab:dataset-overview} summarizes key characteristics.

\begin{table}[t]
\centering
\small
\caption{Overview of \docintent{} and \intentbench{}.}
\label{tab:dataset-overview}
\setlength{\tabcolsep}{4pt}
\renewcommand{\arraystretch}{1.25}
\begin{tabular}{@{}l c c@{}}
\toprule
\textbf{Statistic} & \textbf{\docintent{}} & \textbf{\intentbench{}} \\
\midrule
\# Documents               & 1{,}659       & 216 \\
Total intent cues       & 5{,}641       & 758 \\
Avg.\ cues per doc      & 3.40          & 3.03 \\
Avg.\ tokens per doc    & 157.51        & 175.79 \\
Cue types covered       & 10            & 9 \\
Digital (\%)            & 74.8          & 88.4 \\
Scanned (\%)            & 25.2          & 11.6 \\
\# Document types          & 33            & 18 \\
\bottomrule
\end{tabular}
\end{table}

\subsection{\docintent{}: Training Dataset}

\docintent{} is a multimodal dataset of 1,659 financial documents (invoices, statements, and structured forms) augmented with 5,641 visual intent cues spanning all taxonomy categories. The base documents are drawn from a balanced mix of real-world and synthetic sources (Appendix~\ref{app:annotation-protocol}), so that real documents contribute the layout diversity of production financial workflows while synthetic documents allow deterministic placement of intent cues at specific structural granularities. The 74.8\% digitally rendered / 25.2\% physically annotated and scanned split in Table~\ref{tab:dataset-overview} refers to how the \emph{intent cues} are rendered (simulated overlays versus physical handwriting), not to the source of the underlying base documents; we make this distinction explicit in Appendix~\ref{app:annotation-protocol}. Table~\ref{tab:dataset-overview} summarizes key statistics; per-cue-type breakdowns are in Appendix~\ref{app:datasets}. See Appendix~\ref{app:annotation-protocol} for full annotation protocol details.

% \paragraph{Construction.}
% Base documents are generated from HTML financial templates, with digital cues programmatically injected at controlled structural levels. Handwritten cues are added by annotators and verified for correctness. See Appendix~\ref{app:annotation-protocol} for full annotation protocol details.

\paragraph{Supervision Signal.}
Each document in \docintent{} is paired with a layout-aware intent-resolved transcription that applies all visual cues while preserving two-dimensional document structure (table boundaries, row/column associations, reading order).

\subsection{\intentbench{}: Evaluation Benchmark}

\intentbench{} is a held-out benchmark constructed independently of \docintent{} and excluded entirely from training. It features diverse financial documents with substantial variation in handwriting styles, annotation density, and layouts, testing generalization beyond the training distribution. Ground-truth annotations are intent-resolved, layout-aware transcriptions. See Table~\ref{tab:dataset-overview} for summary statistics and Appendix~\ref{app:datasets} for detailed breakdowns.
\paragraph{Inter-Annotator Agreement.}
Two annotators independently annotated all 216 \intentbench{} documents. Agreement, measured by normalized Levenshtein ratio~\cite{levenshtein1966binary}, yielded a mean similarity of 0.77 ($\sigma=0.12$), indicating moderate-to-strong consistency. Full IAA statistics are in Appendix~\ref{app:iaa}.

\section{Experimentation}

We fine-tune 3B and 7B variants of Qwen2.5-VL-Instruct models on \docintent{} using an 80/20 train/validation split. The fine-tuned models are denoted \texttt{DocIntentOCR-3B} and \texttt{DocIntentOCR-7B}, and evaluated on \intentbench{}. Both models were fine-tuned using LoRA ($r{=}16$, $\alpha{=}32$) on language model layers with the last vision encoder block unfrozen, optimized with AdamW ($1{\times}10^{-4}$/$1{\times}10^{-5}$), for 7 and 5 epochs respectively on two NVIDIA A100 40GB GPUs. To fairly evaluate zero-shot models, we provide a prompt that explicitly defines annotation semantics (Appendix~\ref{app:prompts}), as well as the hyperparameter configurations (Appendix~\ref{app:hyperparameters}).
% fine-tuned models use the training prompt.

\section{Results}

\begin{table}[t]
\centering
\caption{IntentBench OCR Performance}
\label{tab:results}
\resizebox{\columnwidth}{!}{%
\begin{tabular}{lcc}
\hline
\textbf{Model} & \textbf{F1} & \textbf{WER} \\
\hline
Gemini 2.5 Flash & 0.6653 & 1.3306 \\
Gemini 2.5 Pro & 0.6351 & 0.6009 \\
GPT-4.1 & 0.6261 & 0.9058 \\
Claude Sonnet 4.6 & 0.5974 & 0.9308 \\
Mistral-Large-3-675B & 0.6077 & 0.9840 \\
Llama-3.2-90B-Vision & 0.4982 & 1.0986 \\
Qwen3-VL-32B-Instruct & 0.5414 & 1.2720 \\
o3 & 0.7686 & 0.4869 \\
GPT-4o & 0.8052 & 0.4232 \\
Qwen2.5-VL-7B-Instruct (base) & 0.6485 & 0.8172 \\
Qwen2.5-VL-3B-Instruct (base) & 0.6563 & 0.7404 \\
DocIntentOCR-3B & 0.8102 & 0.5622 \\
\textbf{DocIntentOCR-7B} & \textbf{0.8328} & \textbf{0.2589} \\
\hline
\end{tabular}%
}
\end{table}

% Table~\ref{tab:results} reports IntentBench F1 scores. DocIntentOCR-7B achieves 0.833 F1, outperforming GPT-4.1 (0.653) by 18 F1 points. DocIntentOCR-3B (0.810) similarly surpasses all zero-shot baselines, including the much larger Qwen3-VL-32B (0.485), highlighting the importance of intent-aware supervision over model scale.

% Table~\ref{tab:results} reports \intentbench{} F1 scores. The results reveal a clear pattern: intent-aware fine-tuning on \docintent{} is the dominant factor in performance, not model scale or general-purpose capability. Among zero-shot baselines, GPT-4.1 leads at 0.653, while larger open-source models---Qwen3-VL-32B (0.485) and Llama-3.2-90B (0.498)---perform substantially worse, as models trained on clean documents have no mechanism to interpret visual correction marks as semantic edit operations.
% Fine-tuning on \docintent{} yields dramatic gains: \docintentocr{}-3B (0.810) and \docintentocr{}-7B (0.833) both outperform GPT-4.1 (0.653) and all zero-shot baselines. This is practically significant---financial document processing demands privacy and cost constraints that preclude large frontier APIs, and our results show that a compact, locally deployable model trained on intent-aware supervision is a realistic alternative.

Table~\ref{tab:results} reports \intentbench{} F1 scores. The results reveal a clear pattern: intent-aware fine-tuning on \docintent{} is the dominant factor, not model scale. Zero-shot models, including GPT-4.1 (0.626), struggle because they have no mechanism to interpret visual correction marks as semantic edit operations. \docintentocr{}-3B (0.810) and \docintentocr{}-7B (0.833) both outperform all baselines, demonstrating that a compact, locally deployable model trained on intent-aware supervision is a practical and privacy-friendly alternative to large frontier APIs (see Figure~\ref{fig:intent_ocr} for a qualitative comparison with GPT-4.1).

\paragraph{Ablations.} Table~\ref{tab:ablations} isolates the contribution of intent-aware supervision from that of model scale, using the pre-fine-tuning Qwen2.5-VL-Instruct checkpoints as the unsupervised reference under matched architecture, prompt, and decoding settings. Intent supervision yields a +0.155 F1 gain at 3B and a +0.184 F1 gain at 7B, whereas increasing model scale alone (3B$\to$7B, both intent-supervised) yields only +0.023 F1 (roughly 7$\times$ smaller). This confirms that intent-aware supervision, not parameter count, is the dominant driver of performance on this task.

\begin{table}[t]
\centering
\small
\caption{Ablating intent supervision vs.\ model scale on \intentbench{} F1.}
\label{tab:ablations}
\setlength{\tabcolsep}{3pt}
\begin{tabular}{@{}l c c c@{}}
\toprule
\textbf{Ablation} & \textbf{From} & \textbf{To} & \textbf{$\Delta$F1} \\
\midrule
Intent supervision (3B) & 0.656 & 0.810 & +0.155 \\
Intent supervision (7B) & 0.649 & 0.833 & +0.184 \\
Model scale (supervised) & 0.810 & 0.833 & +0.023 \\
\bottomrule
\end{tabular}
\end{table}

\paragraph{Stratified Error Analysis.} Token-level F1 alone conflates distinct failure modes: incorrect intent interpretation in edited regions versus spurious hallucination in unedited ones, and performance can vary substantially across cue types. We decompose performance along both axes in Appendix~\ref{app:stratified}: (i)~per-cue-type Edit Exact Match, and (ii)~edited-region vs.\ non-edited-region F1. \docintentocr{}-7B achieves the highest edit exact match on strikethrough-with-replacement, the most frequent cue type (209 instances, 185 documents), and both fine-tuned variants lead or are competitive across nearly all cue types, with the exception of full-table deletion and arrow-indicated replacement, where \texttt{o3} has a marginal edge. \docintentocr{}-3B matches GPT-4o on edited-region F1 (0.862 vs.\ 0.868) and achieves the highest non-edited-region F1 (0.832) of all evaluated models, indicating it hallucinates least in unannotated content.

\paragraph{Cross-Domain Generalization.} \docintent{} and \intentbench{} are grounded in financial documents. To probe whether intent-aware supervision transfers beyond this domain, we conducted a preliminary cross-domain pilot study on 30 documents spanning three non-financial domains (legal contracts, healthcare records, and real estate contracts; 10 each), evaluating the same models on IntentBench-style annotation interpretation. Full results are reported in Appendix~\ref{app:crossdomain}. \docintentocr{}-3B and \docintentocr{}-7B retain their performance advantage over zero-shot VLMs even out-of-domain (F1 of 0.896 and 0.939, respectively, versus 0.875 for the best zero-shot baseline, \texttt{o3}), and the relative ranking of models is largely preserved across domains, suggesting intent-aware supervision generalizes to the annotation-interpretation skill rather than overfitting to financial-specific content. We emphasize this is a preliminary pilot (30 documents, not a full benchmark); a dedicated cross-domain benchmark is left to future work, and we discuss this limitation further in the Limitations section.
\section{Conclusion and Future Work}

We present intent-aware OCR, a novel document understanding task that interprets human annotations-strikethroughs, arrows, carets, and insertions as semantic edit operations. To support this task, we introduce \docintent{}, a dataset of 1,600+ annotated financial documents, and \intentbench{}, a 216-document real-world benchmark. Fine-tuned on \docintent{}, \docintentocr{} in both 3B and 7B configurations substantially outperform GPT-4.1 and other frontier models in the zero-shot setting, demonstrating that intent-aware supervision is the critical factor rather than model scale. A preliminary cross-domain pilot study (Appendix~\ref{app:crossdomain}) further suggests this advantage extends beyond financial documents to legal, healthcare, and real estate contracts. Future work includes releasing a dedicated cross-domain benchmark with broader document coverage, expanding the taxonomy to handle compositional and ambiguous annotations (e.g., circling, underlining, marginal comments), and scaling training data via synthetic image generation.

\section*{Limitations}
Our main dataset and evaluation are grounded in financial documents; while our cross-domain pilot study (Appendix~\ref{app:crossdomain}, 30 documents across legal, healthcare, and real estate domains) offers preliminary evidence of generalization, it is not a substitute for a full cross-domain benchmark, and broader validation remains future work. The taxonomy does not cover all annotation styles such as circling, underlining, and marginal comments; these were excluded because they are structurally ambiguous (e.g., a circle can indicate deletion, emphasis, or a query depending on context) and admit no operationally unambiguous definition, so undefined annotation types are currently treated as a no-op by the model. Extending the taxonomy to these ambiguous conventions, and to other domains' annotation styles (e.g., accept/reject semantics in legal redlining), is an important direction. The predominantly digital training data (74.8\%) may limit robustness to diverse real-world handwriting. Inter-annotator agreement of 0.77 indicates that spatially ambiguous annotations admit subjective interpretations; manual inspection of low-agreement pairs suggests spatially ambiguous annotation placement (e.g., a caret positioned between two candidate insertion points, or a correction near a cell boundary) is the primary source of disagreement, rather than systematic differences in annotator understanding of the taxonomy. Our token-level F1 and WER metrics, while complemented by per-cue-type and edited/non-edited-region analyses (Appendix~\ref{app:stratified}), do not yet provide a full qualitative taxonomy of systematic vs.\ random failure modes. Finally, the structural scopes in our taxonomy (token, field, cell, row, column, table) are designed primarily for tabular layouts; documents with irregular or free-form layouts may require alternative scope definitions.

\section*{Ethics Statement}
Documents in \docintent{} and \intentbench{} are either synthetically generated or sourced from publicly available document collections and templates (Appendix~\ref{app:annotation-protocol}); no private, non-public financial records or personally identifiable information were used. Human annotators were recruited voluntarily, informed of the task purpose, and compensated fairly for their time. Annotation instructions were designed to avoid subjective or harmful content.

Portions of this paper were drafted with the assistance of large language model tools (Claude, OpenAI). All scientific content, experimental results, and conclusions were produced and verified by the authors.

% \section{ Summary, conclusion, and future work}
% We present intent-aware OCR, a novel document understanding task that interprets human annotations—strikethroughs, arrows, carets, and insertions—as semantic edit operations. To support this task, we introduce \docintent{}, a dataset of 1,600+ annotated financial documents, and \intentbench{}, a 250-document real-world benchmark. Fine-tuned on \docintent{}, both our 3B and 7B models substantially outperform GPT-4.1 and other frontier models in the zero-shot setting, demonstrating that intent-aware supervision is the critical factor rather than model scale. Future work includes validating the formulation on legal and medical documents, expanding the taxonomy to handle compositional annotations, and scaling training data via synthetic image generation.

% \paragraph{Limitations}
% Our dataset and evaluation are restricted to financial documents, and generalization to other domains remains unvalidated. The taxonomy does not cover all annotation styles such as circling, underlining, and marginal comments, and the predominantly digital training data (74.8\%) may limit robustness to diverse real-world handwriting. Inter-annotator agreement of 0.77 indicates that spatially ambiguous annotations admit subjective interpretations.

% ---- Bibliography ----
% \bibliographystyle is already set by acl.sty; calling it again here
% causes BibTeX to fail with "Illegal, another \bibstyle command".
\bibliography{main}
\appendix
% -----------------------------------------------------------------
\section{Intent Cue Taxonomy}
% The taxonomy focuses on financial documents, but the three edit operation categories (deletion, replacement, insertion) are domain-agnostic primitives, and the visual cues that express them (strikethroughs, carets, arrows) are conventional marks used across professional domains. Extending coverage to legal redlining, medical records, or engineering change orders may require additional cue types and scope definitions; empirical validation in non-financial domains remains an important direction for future work.

Table~\ref{tab:intent-taxonomy} enumerates the full taxonomy. While the ten cue types are grounded in financial documents, the three edit operation categories (deletion, replacement, insertion) are domain-agnostic primitives, and the visual cues that express them (strikethroughs, carets, arrows) are conventional marks used across professional domains. Extending coverage to legal redlining, medical records, or engineering change orders may require additional cue types and scope definitions; empirical validation in non-financial domains remains an important direction for future work.
\label{app:taxonomy}

\begin{table*}[t]
\centering
\small
\caption{Taxonomy of visual and semantic intent cues for intent-aware OCR. Each cue induces a deterministic edit over the base transcription at a specific structural granularity.}
\label{tab:intent-taxonomy}
\setlength{\tabcolsep}{3pt}
\renewcommand{\arraystretch}{1.25}
\resizebox{\textwidth}{!}{%
\begin{tabular}{@{}l l l p{5.8cm}@{}}
\toprule
\textbf{Category} & \textbf{Cue Type} & \textbf{Scope} & \textbf{Expected OCR Behavior} \\
\midrule
\multirow{4}{*}{Deletion}
& Strikethrough          & Token / Field  & Marked content is omitted from the output. \\
& Row Strikethrough      & Table Row      & All cells in the struck-through row are removed. \\
& Column Strikethrough   & Table Column   & All values in the struck-through column are removed. \\
& Full-Table Strikethrough & Table        & The entire table is omitted from the output. \\
\addlinespace
\midrule
\multirow{5}{*}{Replacement}
& Strikethrough + Handwritten Value & Token / Field & Struck-through text is replaced by the nearby handwritten value. \\
& Arrow-Indicated Replacement       & Token / Field & Struck-through text is replaced by the value the arrow points to. \\
& Cell Correction                   & Table Cell    & The cell value is replaced by the handwritten correction. \\
& Row Correction                    & Table Row     & Each cell in the row is substituted with its per-cell replacement. \\
& Column Correction                 & Table Column  & Each cell in the column is substituted with its replacement value. \\
\addlinespace
\midrule
\multirow{3}{*}{Insertion}
& Caret Insertion & Token / Field & New handwritten text is inserted at the caret position without deleting existing content. \\
& Arrow-Indicated Addition  & Token / Field & New content is inserted at the location indicated by an arrow. \\
& Inline Handwritten Addition & Token / Field & New handwritten content is inserted adjacent to existing text without explicit caret. \\
\addlinespace
\midrule
\multirow{3}{*}{Selection / Validation}
& Tick Mark (\checkmark) & Token / Field & Marked content is explicitly selected or confirmed as valid. \\
& Cross Mark (X)         & Token / Field & Marked content is explicitly rejected or invalidated (often implying deletion). \\
& Filled Circle / Highlight & Token / Field & Circled or highlighted content is emphasized as relevant or selected. \\
\addlinespace
\midrule
\multirow{3}{*}{Natural Language}
& Instructional Override & Variable & Free-form text (e.g., “replace with 500”) explicitly overrides existing content. \\
& Instructional Deletion & Variable & Textual cues (e.g., “ignore this row”) remove specified content from output. \\
& Instructional Addition & Variable & Textual cues (e.g., “add GST”) introduce new content at contextually appropriate locations. \\
\bottomrule
\end{tabular}%
}
\end{table*}

\begin{figure*}[p]
\centering
\begin{subfigure}[t]{0.46\textwidth}
\centering
\includegraphics[width=\linewidth,height=1.0cm,keepaspectratio]{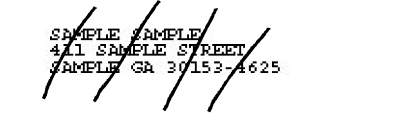}\\[1pt]
{\footnotesize\textbf{Strikethrough w/o Replacement} (Deletion).}
\end{subfigure}
\hfill
\begin{subfigure}[t]{0.46\textwidth}
\centering
\begin{minipage}[c][1.0cm][c]{\linewidth}
\centering
\footnotesize(empty)
\end{minipage}\\[1pt]
{\footnotesize Intent-resolved OCR output.}
\end{subfigure}
\label{fig:cue01}

\vspace{0.2em}

\noindent\rule{\textwidth}{0.4pt}

\vspace{0.2em}
\begin{subfigure}[t]{0.46\textwidth}
\centering
\includegraphics[width=\linewidth,height=1.9cm,keepaspectratio]{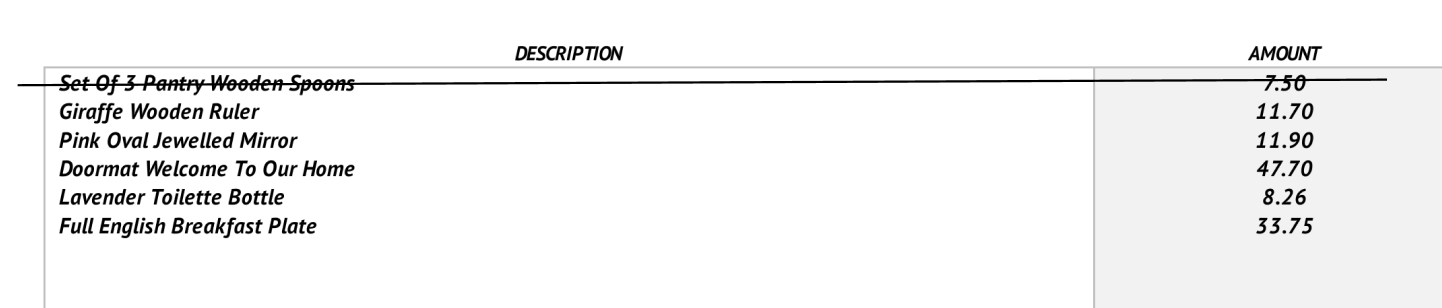}\\[1pt]
{\footnotesize\textbf{Table Row Deletion} (Deletion).}
\end{subfigure}
\hfill
\begin{subfigure}[t]{0.46\textwidth}
\centering
\begin{minipage}[c][1.9cm][c]{\linewidth}
\centering
\scriptsize
\begin{tabular}{@{}l r@{}}
Description & Amount\\
\midrule
Giraffe Wooden Ruler & 11.70\\
Pink Oval Jewelled Mirror & 11.90\\
Doormat Welcome To Our Home & 47.70\\
Lavender Toilette Bottle & 8.26\\
Full English Breakfast Plate & 33.75\\
\end{tabular}
\end{minipage}\\[1pt]
{\footnotesize Intent-resolved OCR output.}
\end{subfigure}
\label{fig:cue02}

\vspace{0.2em}

\noindent\rule{\textwidth}{0.4pt}

\vspace{0.2em}
\begin{subfigure}[t]{0.46\textwidth}
\centering
\includegraphics[width=\linewidth,height=2.4cm,keepaspectratio]{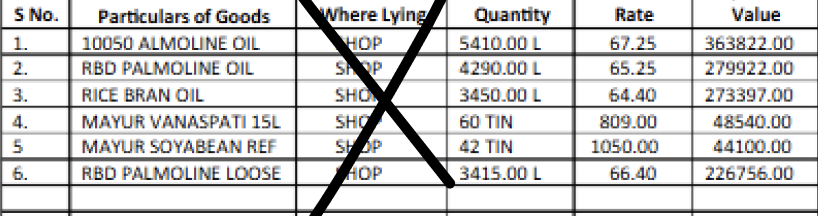}\\[1pt]
{\footnotesize\textbf{Table Column Deletion} (Deletion).}
\end{subfigure}
\hfill
\begin{subfigure}[t]{0.46\textwidth}
\centering
\begin{minipage}[c][2.4cm][c]{\linewidth}
\centering
\scriptsize
\begin{tabular}{@{}l r r r@{}}
S No. / Particulars of Goods & Quantity & Rate & Value\\
\midrule
1. 10050 ALMOLINE OIL & 5410.00 L & 67.25 & 363822.00\\
2. RBD PALMOLINE OIL & 4290.00 L & 65.25 & 279922.00\\
3. RICE BRAN OIL & 3450.00 L & 64.40 & 273397.00\\
4. MAYUR VANASPATI 15L & 60 TIN & 809.00 & 48540.00\\
5. MAYUR SOYABEAN REF & 42 TIN & 1050.00 & 44100.00\\
6. RBD PALMOLINE LOOSE & 3415.00 L & 66.40 & 226756.00\\
\end{tabular}
\end{minipage}\\[1pt]
{\footnotesize Intent-resolved OCR output.}
\end{subfigure}
\label{fig:cue03}

\vspace{0.2em}

\noindent\rule{\textwidth}{0.4pt}

\vspace{0.2em}
\begin{subfigure}[t]{0.46\textwidth}
\centering
\includegraphics[width=\linewidth,height=1.0cm,keepaspectratio]{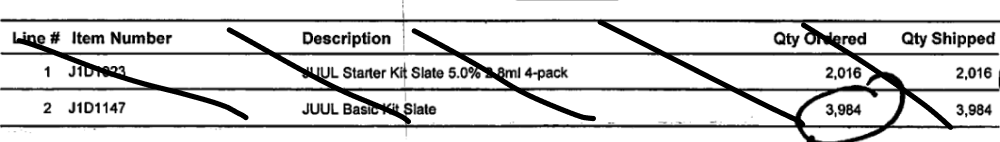}\\[1pt]
{\footnotesize\textbf{Full Table Deletion} (Deletion).}
\end{subfigure}
\hfill
\begin{subfigure}[t]{0.46\textwidth}
\centering
\begin{minipage}[c][1.0cm][c]{\linewidth}
\centering
\footnotesize(empty)
\end{minipage}\\[1pt]
{\footnotesize Intent-resolved OCR output.}
\end{subfigure}
\label{fig:cue04}

\vspace{0.2em}

\noindent\rule{\textwidth}{0.4pt}

\vspace{0.2em}
\begin{subfigure}[t]{0.46\textwidth}
\centering
\includegraphics[width=\linewidth,height=1.0cm,keepaspectratio]{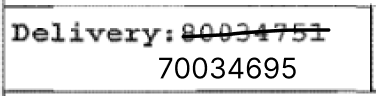}\\[1pt]
{\footnotesize\textbf{Strikethrough w/ Replacement} (Replacement).}
\end{subfigure}
\hfill
\begin{subfigure}[t]{0.46\textwidth}
\centering
\begin{minipage}[c][1.0cm][c]{\linewidth}
\centering
\footnotesize Delivery: 70034695
\end{minipage}\\[1pt]
{\footnotesize Intent-resolved OCR output.}
\end{subfigure}
\label{fig:cue05}

\vspace{0.2em}

\noindent\rule{\textwidth}{0.4pt}

\vspace{0.2em}
\begin{subfigure}[t]{0.46\textwidth}
\centering
\includegraphics[width=\linewidth,height=1.0cm,keepaspectratio]{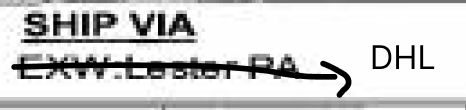}\\[1pt]
{\footnotesize\textbf{Arrow-Indicated Replacement} (Replacement).}
\end{subfigure}
\hfill
\begin{subfigure}[t]{0.46\textwidth}
\centering
\begin{minipage}[c][1.0cm][c]{\linewidth}
\centering
\footnotesize SHIP VIA\\ DHL
\end{minipage}\\[1pt]
{\footnotesize Intent-resolved OCR output.}
\end{subfigure}
\label{fig:cue06}

\vspace{0.2em}

\noindent\rule{\textwidth}{0.4pt}

\vspace{0.2em}
\begin{subfigure}[t]{0.46\textwidth}
\centering
\includegraphics[width=\linewidth,height=1.3cm,keepaspectratio]{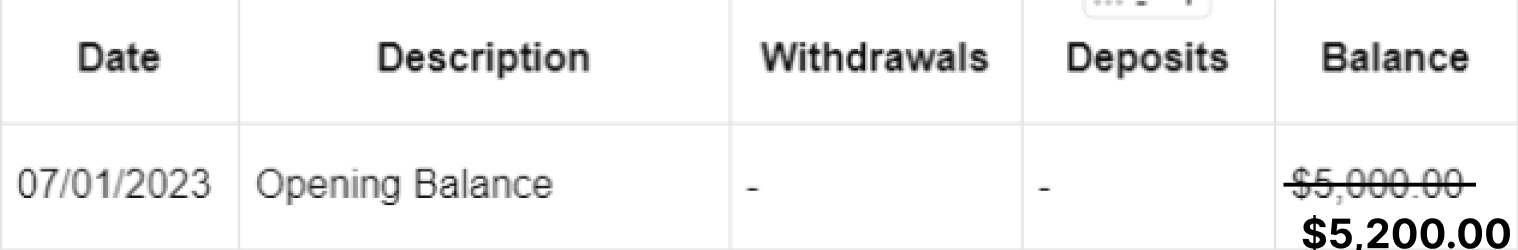}\\[1pt]
{\footnotesize\textbf{Table Cell Replacement} (Replacement).}
\end{subfigure}
\hfill
\begin{subfigure}[t]{0.46\textwidth}
\centering
\begin{minipage}[c][1.3cm][c]{\linewidth}
\centering
\scriptsize
\begin{tabular}{@{}l l c c r@{}}
Date & Description & Withdrawals & Deposits & Balance\\
\midrule
07/01/2023 & Opening Balance & - & - & \$5,200.00\\
\end{tabular}
\end{minipage}\\[1pt]
{\footnotesize Intent-resolved OCR output.}
\end{subfigure}
\label{fig:cue07}

\vspace{0.2em}

\noindent\rule{\textwidth}{0.4pt}

\vspace{0.2em}
\begin{subfigure}[t]{0.46\textwidth}
\centering
\includegraphics[width=\linewidth,height=1.7cm,keepaspectratio]{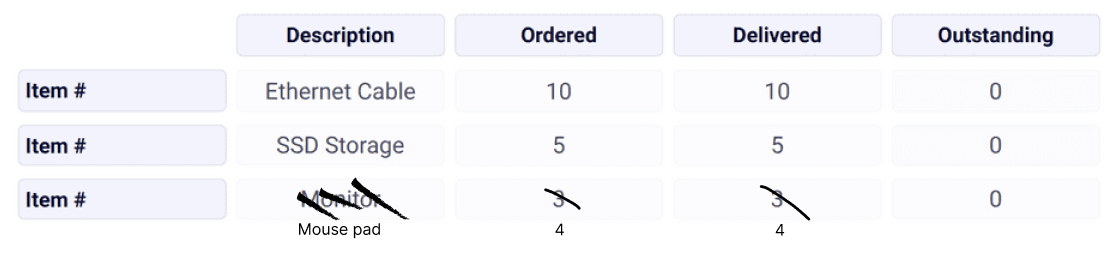}\\[1pt]
{\footnotesize\textbf{Table Row Replacement} (Replacement).}
\end{subfigure}
\hfill
\begin{subfigure}[t]{0.46\textwidth}
\centering
\begin{minipage}[c][1.7cm][c]{\linewidth}
\centering
\scriptsize
\begin{tabular}{@{}l l c c c@{}}
& Description & Ordered & Delivered & Outstanding\\
\midrule
Item \# & Ethernet Cable & 10 & 10 & 0\\
Item \# & SSD Storage & 5 & 5 & 0\\
Item \# & Mouse pad & 4 & 4 & 0\\
\end{tabular}
\end{minipage}\\[1pt]
{\footnotesize Intent-resolved OCR output.}
\end{subfigure}
\label{fig:cue08}

\vspace{0.2em}

\noindent\rule{\textwidth}{0.4pt}

\vspace{0.2em}
\begin{subfigure}[t]{0.46\textwidth}
\centering
\includegraphics[width=\linewidth,height=2.0cm,keepaspectratio]{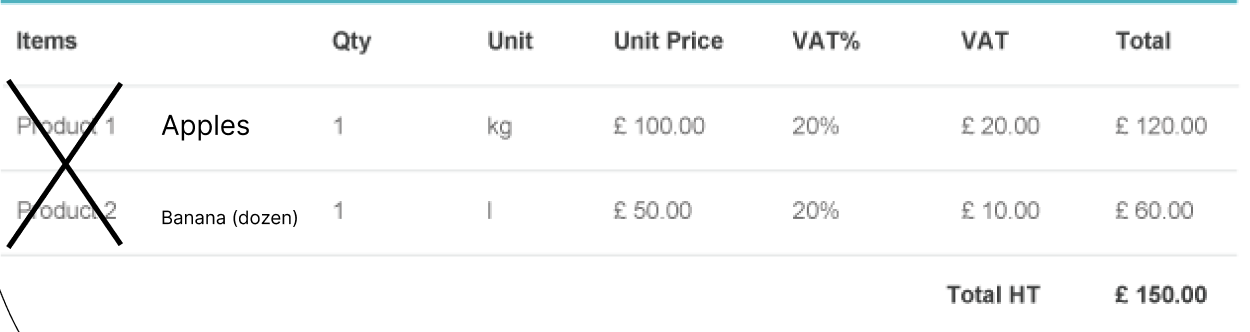}\\[1pt]
{\footnotesize\textbf{Table Column Replacement} (Replacement).}
\end{subfigure}
\hfill
\begin{subfigure}[t]{0.46\textwidth}
\centering
\begin{minipage}[c][2.0cm][c]{\linewidth}
\centering
\scriptsize
\begin{tabular}{@{}l c c r r r@{}}
Items & Qty & Unit & Unit Price & VAT\% & Total\\
\midrule
Apples & 1 & kg & \pounds100.00 & 20\% & \pounds120.00\\
Banana (dozen) & 1 & l & \pounds50.00 & 20\% & \pounds60.00\\
\end{tabular}\\[2pt]
Total HT: \pounds150.00
\end{minipage}\\[1pt]
{\footnotesize Intent-resolved OCR output.}
\end{subfigure}
\label{fig:cue09}

\vspace{0.2em}

\noindent\rule{\textwidth}{0.4pt}

\vspace{0.2em}
\begin{subfigure}[t]{0.46\textwidth}
\centering
\includegraphics[width=\linewidth,height=1.0cm,keepaspectratio]{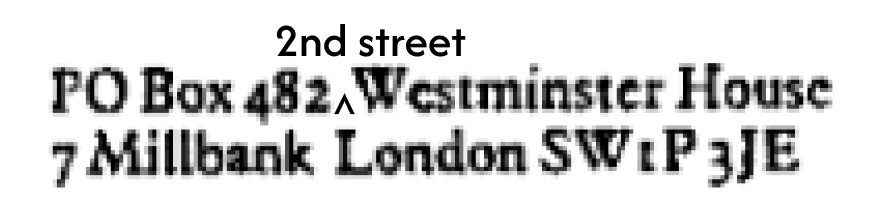}\\[1pt]
{\footnotesize\textbf{Caret Insertion} (Insertion).}
\end{subfigure}
\hfill
\begin{subfigure}[t]{0.46\textwidth}
\centering
\begin{minipage}[c][1.0cm][c]{\linewidth}
\centering
\footnotesize PO Box 482, 2nd Street, Westminster House\\ 7 Millbank London SW1P 3JE
\end{minipage}\\[1pt]
{\footnotesize Intent-resolved OCR output.}
\end{subfigure}
\label{fig:cue10}

\caption{Representative annotated examples of all ten intent cue types in \docintent{}/\intentbench{}, shown alongside the literal intent-resolved OCR output.}
\label{fig:cue-gallery}
\end{figure*}

Figure~\ref{fig:cue-gallery} shows one representative annotated example per cue type actually present in \docintent{} and \intentbench{} (the 10 cue types summarized in Table~\ref{tab:dataset-overview}).

\clearpage

\clearpage

\section{Dataset Statistics}
\label{app:datasets}
Tables~\ref{tab:cue-distribution} and~\ref{tab:cue-distribution-intentbench} report the per-cue-type distribution of visual intent cues in \docintent{} and \intentbench{}, respectively. Replacement cues dominate both splits, with strikethrough-with-replacement and arrow-indicated replacement being the most frequent types.
  
\begin{table}[h]
\centering
\small
\caption{Distribution of visual intent cues in \docintent{}. Each row reports the total number of cue instances, the number of documents containing at least one instance, and the average frequency per document across all 1{,}659 documents.}
\label{tab:cue-distribution}
\setlength{\tabcolsep}{2pt}
\renewcommand{\arraystretch}{1.25}
\resizebox{0.48\textwidth}{!}{%
\begin{tabular}{@{}l l r r r@{}}
\toprule
\textbf{Category} & \textbf{Cue Type} & \textbf{Total} & \textbf{Docs} & \textbf{Avg./Doc} \\
\midrule
\multirow{4}{*}{Deletion}
& Strikethrough w/o Replacement  & 986   & 833   & 0.59 \\
& Table Row Deletion             & 189   & 169   & 0.11 \\
& Table Column Deletion          & 333   & 303   & 0.20 \\
& Full Table Deletion            & 133   & 130   & 0.08 \\
\addlinespace
\midrule
\multirow{5}{*}{Replacement}
& Strikethrough w/ Replacement   & 1{,}880 & 1{,}233 & 1.13 \\
& Arrow-Indicated Replacement    & 1{,}214 & 1{,}010 & 0.73 \\
& Table Cell Replacement         & 463   & 315   & 0.28 \\
& Table Row Replacement          & 16    & 12    & 0.01 \\
& Table Column Replacement       & 45    & 33    & 0.03 \\
\addlinespace
\midrule
Insertion
& Caret Insertion                & 382   & 331   & 0.23 \\
\midrule
\multicolumn{2}{@{}l}{\textbf{Total}} & \textbf{5{,}641} & -- & \textbf{3.40} \\
\bottomrule
\end{tabular}%
}
\end{table}

\begin{table}[h]
\centering
\small
\caption{Distribution of visual intent cues in \intentbench{}. Each row reports the total number of cue instances, the number of documents containing at least one instance, and the average frequency per document.}
\label{tab:cue-distribution-intentbench}
\setlength{\tabcolsep}{2pt}
\renewcommand{\arraystretch}{1.25}
\resizebox{0.48\textwidth}{!}{%
\begin{tabular}{@{}l l r r r@{}}
\toprule
\textbf{Category} & \textbf{Cue Type} & \textbf{Total} & \textbf{Docs} & \textbf{Avg./Doc} \\
\midrule
\multirow{4}{*}{Deletion}
& Strikethrough w/o Replacement  & 160   & 143   & 0.74 \\
& Table Row Deletion             & 1     & 1     & 0.00 \\
& Table Column Deletion          & 30    & 30    & 0.14 \\
& Full Table Deletion            & 12    & 12    & 0.06 \\
\addlinespace
\midrule
\multirow{4}{*}{Replacement}
& Strikethrough w/ Replacement   & 209   & 185   & 0.97 \\
& Arrow-Indicated Replacement    & 192   & 180   & 0.89 \\
& Table Cell Replacement         & 28    & 23    & 0.13 \\
& Table Column Replacement       & 1     & 1     & 0.00 \\
\addlinespace
\midrule
Insertion
& Caret Insertion                & 22    & 21    & 0.10 \\
\midrule
\multicolumn{2}{@{}l}{\textbf{Total}} & \textbf{655} & -- & \textbf{3.03} \\
\bottomrule
\end{tabular}%
    }
\end{table}

% -----------------------------------------------------------------

\section{Annotation Protocol}
\label{app:annotation-protocol}

\paragraph{Data Sources.} \docintent{} base documents (prior to annotation) are drawn from a balanced mix of real-world and synthetic sources spanning 33 document types. \textbf{Real-world documents} (approximately 53\% of the corpus) are sourced from the publicly available UCSF Industry Documents Library~\cite{ucsf_industry_documents} and other publicly distributed finance document collections, contributing the layout and content diversity found in production financial workflows. \textbf{Synthetic documents} (approximately 47\%) are produced by two complementary pipelines: (i)~custom-built HTML financial templates populated with synthetically generated content (names, line items, amounts) designed to be mathematically consistent (e.g., correct subtotals, taxes, and grand totals) and rendered to images via a headless browser, and (ii)~the Inv3D dataset~\cite{hertlein2023inv3d}, which provides high-resolution synthetic 3D-rendered invoices. Synthetic documents give precise control over layout, table structure, and spatial organization (properties essential for deterministic placement of intent cues at specific structural granularities), while the real-world portion grounds the dataset in authentic document diversity. We note that the 74.8\%~digital / 25.2\%~scanned split reported in Table~\ref{tab:dataset-overview} refers to the rendering of the \emph{intent cues} (digitally simulated annotations vs. physically handwritten and scanned annotations) overlaid on these base documents, not to the source of the base documents themselves.

Handwritten annotations were produced by five annotators with academic backgrounds in commerce and prior exposure to accounting and business documentation. Annotators were given document-specific correction instructions specifying which content to delete, replace, or insert, and were asked to apply corrections using natural handwriting without access to model predictions or automated suggestions. Each annotated document was independently reviewed by a second annotator to verify that all intended cues were correctly applied and visually unambiguous. Annotators were fairly compensated, and the study adhered to established ethical standards.

% -----------------------------------------------------------------

\section{Inter-Annotator Agreement}
\label{app:iaa}

To assess the reliability of \intentbench{} reference annotations, two annotators independently produced intent-resolved transcriptions for all 216 documents; agreement was quantified using the normalized Levenshtein ratio~\cite{levenshtein1966binary}. Table~\ref{tab:iaa} reports the resulting similarity statistics, with a mean score of 0.77 indicating moderate-to-strong consistency across annotators. 

\begin{table}[!htb]
\centering
% \small
\caption{Inter-Annotator Agreement (IAA) on \intentbench{} annotations using fuzzy string similarity (normalized Levenshtein ratio).}
\label{tab:iaa}
\setlength{\tabcolsep}{4pt}
\renewcommand{\arraystretch}{1.25}
\begin{tabular}{@{}l c@{}}
\toprule
\textbf{Metric} & \textbf{Value} \\
\midrule
Number of annotation pairs & 216 \\
Mean similarity            & 0.77 \\
Minimum similarity         & 0.31 \\
Maximum similarity         & 0.99 \\
Standard deviation         & 0.12 \\
\bottomrule
\end{tabular}
\end{table}

% -----------------------------------------------------------------
\section{Experimental Configurations}
\label{app:experiment_config}

\subsection{Prompts}
\label{app:prompts}

To evaluate \texttt{DocIntentOCR-3B} and \texttt{DocIntentOCR-7B}, we use the same instruction as employed during training:
\textit{``You are an OCR assistant. Analyze the document image and transcribe all visible text, interpreting any annotations as edit instructions. Produce the final corrected transcription.''}

For all other non-finetuned models and closed-source models, we use a slightly expanded prompt to explicitly define annotation semantics:
\textit{``You are an OCR assistant for annotated financial documents. Interpret annotations as edit instructions (e.g., strikethroughs for deletions or replacements, arrows for replacements, carets for insertions, etc.). Output the final corrected transcription, preserving layout.''}

\subsection{Inference Hyperparameters}
\label{app:hyperparameters}
All models were evaluated with temperature 0.1, top-$p$ 0.90, and a maximum output length of 8{,}000 tokens. Zero-shot baselines were accessed via the OpenAI API (GPT-4.1), Anthropic API (Claude Sonnet 4.6), and AWS Bedrock (Llama-3.2-90B and Mistral-Large-3-675B).

% ------------------------------------------------------

% \section{Illustrative Example OCR}
% The OCR for the illustrative example

% \subsection{Intent-Aware OCR using \docintentocr{}} 

% \subsection{Frontier-Model OCR}
% \paragraph{GPT-4.1}

% \paragraph{LLaMA-3.2-Vision}

\section{Stratified Error Analysis}
\label{app:stratified}

\subsection{Per-Cue-Type Edit Exact Match}
Table~\ref{tab:eem-cue-type} reports Edit Exact Match (EEM), a stricter metric than token-level F1 that scores 1 only if the correct edited token appears \emph{and} the struck-through original is absent, broken down by cue type, for all evaluated models on \intentbench{}. Cue-type membership is determined from \intentbench{}'s ground-truth annotation flags. We report the seven most frequent cue types (sample sizes for the remaining rare types, e.g., table row/column replacement, are too small for stable per-type estimates; see Table~\ref{tab:cue-distribution-intentbench}).

\begin{table*}[t]
\centering
\small
\caption{Edit Exact Match (EEM) by cue type on \intentbench{}, for the seven most frequent cue types. Bold marks the best score per row.}
\label{tab:eem-cue-type}
\setlength{\tabcolsep}{3pt}
\renewcommand{\arraystretch}{1.2}
\resizebox{\textwidth}{!}{%
\begin{tabular}{@{}l c c c c c c c c c c c c c@{}}
\toprule
\textbf{Cue Type} & \textbf{\docintentocr{}-7B} & \textbf{\docintentocr{}-3B} & \textbf{Qwen2.5-VL-7B} & \textbf{Qwen2.5-VL-3B} & \textbf{GPT-4.1} & \textbf{GPT-4o} & \textbf{o3} & \textbf{Claude-S.-4.6} & \textbf{Gemini-2.5-Flash} & \textbf{Qwen3-VL-32B} & \textbf{Gemini-2.5-Pro} & \textbf{Llama-3.2-90B} & \textbf{Mistral-L.-3} \\
\midrule
Strikethrough w/o Replacement & 0.83 & \textbf{0.85} & 0.67 & 0.79 & 0.73 & 0.83 & 0.81 & 0.71 & 0.81 & 0.67 & 0.83 & 0.77 & 0.73 \\
Table Column Deletion         & \textbf{0.81} & \textbf{0.81} & 0.69 & 0.67 & 0.69 & \textbf{0.81} & 0.71 & 0.70 & 0.72 & 0.59 & 0.69 & 0.44 & 0.69 \\
Full Table Deletion           & 0.68 & 0.71 & 0.71 & 0.68 & 0.72 & \textbf{0.80} & \textbf{0.80} & 0.68 & 0.64 & 0.57 & 0.71 & 0.54 & 0.69 \\
Strikethrough w/ Replacement  & \textbf{0.79} & 0.77 & 0.55 & 0.57 & 0.72 & 0.78 & 0.75 & 0.70 & 0.66 & 0.62 & 0.62 & 0.57 & 0.57 \\
Arrow-Indicated Replacement   & 0.73 & 0.75 & 0.53 & 0.44 & 0.66 & 0.76 & \textbf{0.77} & 0.69 & 0.68 & 0.61 & 0.64 & 0.37 & 0.50 \\
Table Cell Replacement        & 0.88 & 0.94 & 0.88 & 0.59 & 0.88 & \textbf{1.00} & \textbf{1.00} & \textbf{1.00} & 0.88 & 0.94 & 0.71 & 0.88 & 0.82 \\
Caret Insertion                & \textbf{1.00} & \textbf{1.00} & 0.80 & 0.80 & \textbf{1.00} & \textbf{1.00} & 0.60 & \textbf{1.00} & \textbf{1.00} & \textbf{1.00} & 0.60 & 0.60 & 0.40 \\
\bottomrule
\end{tabular}%
}
\end{table*}

\docintentocr{}-7B achieves the highest EEM on strikethrough with replacement, the most frequent cue type (209 instances, 185 documents), and \docintentocr{}-3B leads on strikethrough without replacement. Both fine-tuned variants lead on table column deletion. Across all deletion and replacement cue types, the fine-tuned models consistently outperform their own backbone (Qwen2.5-VL-7B/3B) by substantial margins. The fine-tuned models are competitive with or lead all cue types except full-table deletion and arrow-indicated replacement, where \texttt{o3} has a marginal edge (0.80 vs.\ 0.68--0.71 and 0.77 vs.\ 0.73--0.75, respectively).

\subsection{Edited vs.\ Non-Edited Region Analysis}
Token-level F1 alone conflates errors in edited regions (incorrect intent interpretation or edit application) with errors in non-edited regions (spurious hallucination or text degradation). Table~\ref{tab:edited-nonedited} decomposes performance into Edited Region F1 and Non-Edited Region F1, using the known cue bounding boxes from \intentbench{} annotations to segment output tokens into edited vs.\ non-edited spans and computing F1 separately for each.

\begin{table}[t]
\centering
\small
\caption{Edited-region vs.\ non-edited-region F1 on \intentbench{}. Bold marks the best score per column.}
\label{tab:edited-nonedited}
\setlength{\tabcolsep}{4pt}
\renewcommand{\arraystretch}{1.2}
\resizebox{\columnwidth}{!}{%
\begin{tabular}{@{}l c c@{}}
\toprule
\textbf{Model} & \textbf{Edited Region F1} & \textbf{Non-Edited Region F1} \\
\midrule
\docintentocr{}-7B & 0.8624 & 0.8170 \\
\docintentocr{}-3B & 0.8624 & \textbf{0.8321} \\
Qwen2.5-VL-7B-Instruct & 0.7151 & 0.7338 \\
Qwen2.5-VL-3B-Instruct & 0.6699 & 0.7074 \\
GPT-4.1 & 0.8204 & 0.6815 \\
GPT-4o & \textbf{0.8677} & 0.8206 \\
o3 & 0.8613 & 0.8052 \\
Claude Sonnet 4.6 & 0.8280 & 0.6888 \\
Gemini 2.5 Flash & 0.8140 & 0.7145 \\
Qwen3-VL-32B-Instruct & 0.7774 & 0.6041 \\
Gemini 2.5 Pro & 0.7710 & 0.6938 \\
Llama-3.2-90B-Vision & 0.6516 & 0.5601 \\
Mistral-Large-3-675B & 0.7204 & 0.6828 \\
\bottomrule
\end{tabular}%
}
\end{table}

\docintentocr{}-3B matches GPT-4o on edited-region F1 (0.8624 vs.\ 0.8677) and achieves the highest non-edited-region F1 (0.8321) of all evaluated models, indicating it hallucinates least in unannotated content. In contrast, several strong zero-shot baselines (e.g., GPT-4.1, Claude Sonnet 4.6, Qwen3-VL-32B) show a pronounced gap between edited- and non-edited-region F1, suggesting their errors are concentrated in unedited content rather than in misapplied edits.

% -----------------------------------------------------------------
\section{Cross-Domain Pilot Study}
\label{app:crossdomain}

\docintent{} and \intentbench{} are both grounded in financial documents, and the main results do not provide direct evidence of cross-domain transfer. To provide a preliminary signal on generalization, we conducted a focused cross-domain pilot study on 30 documents spanning three non-financial domains: legal contracts (10), healthcare records (10), and real estate contracts (10), evaluating the same set of models on IntentBench-style annotation interpretation (intent cues overlaid on the documents following the same taxonomy and annotation protocol as \intentbench{}). Table~\ref{tab:crossdomain} reports F1 and WER.

\begin{table}[t]
\centering
\small
\caption{Cross-domain pilot study (30 documents: legal, healthcare, real estate). Bold marks the best score per column.}
\label{tab:crossdomain}
\setlength{\tabcolsep}{4pt}
\renewcommand{\arraystretch}{1.2}
\begin{tabular}{@{}l c c@{}}
\toprule
\textbf{Model} & \textbf{F1} & \textbf{WER} \\
\midrule
GPT-4.1 & 0.8006 & 0.4677 \\
Mistral-Large-3-675B & 0.7698 & 0.4578 \\
Gemini 2.5 Flash & 0.8180 & 0.3580 \\
Gemini 2.5 Pro & 0.8520 & 0.2980 \\
Llama-3.2-90B-Vision & 0.8350 & 0.3270 \\
Qwen3-VL-32B-Instruct & 0.7502 & 0.6229 \\
GPT-4o & 0.8661 & 0.2343 \\
o3 & 0.8749 & 0.2240 \\
Qwen2.5-VL-7B-Instruct (base) & 0.7971 & 0.4397 \\
Qwen2.5-VL-3B-Instruct (base) & 0.7664 & 0.4650 \\
\docintentocr{}-3B & 0.8961 & 0.1540 \\
\docintentocr{}-7B & \textbf{0.9389} & \textbf{0.0873} \\
\bottomrule
\end{tabular}
\end{table}

Two observations follow. First, \docintentocr{}-3B and \docintentocr{}-7B maintain their performance advantage over zero-shot VLMs even on out-of-domain documents, suggesting that intent-aware supervision generalizes to the annotation-interpretation task rather than overfitting to financial-specific content. Second, the relative ordering of models is largely preserved across domains, indicating that \intentbench{} rankings are reasonably stable signals rather than domain artifacts. We stress that this is a preliminary pilot (30 documents, not a full benchmark), and we are committed to releasing a dedicated cross-domain benchmark with broader document coverage as future work.

% -----------------------------------------------------------------

\section{Output comparison of DocIntentOCR-7B with GPT-4.1}

\begin{figure*}[t]
    \centering
\includegraphics[width=\textwidth]{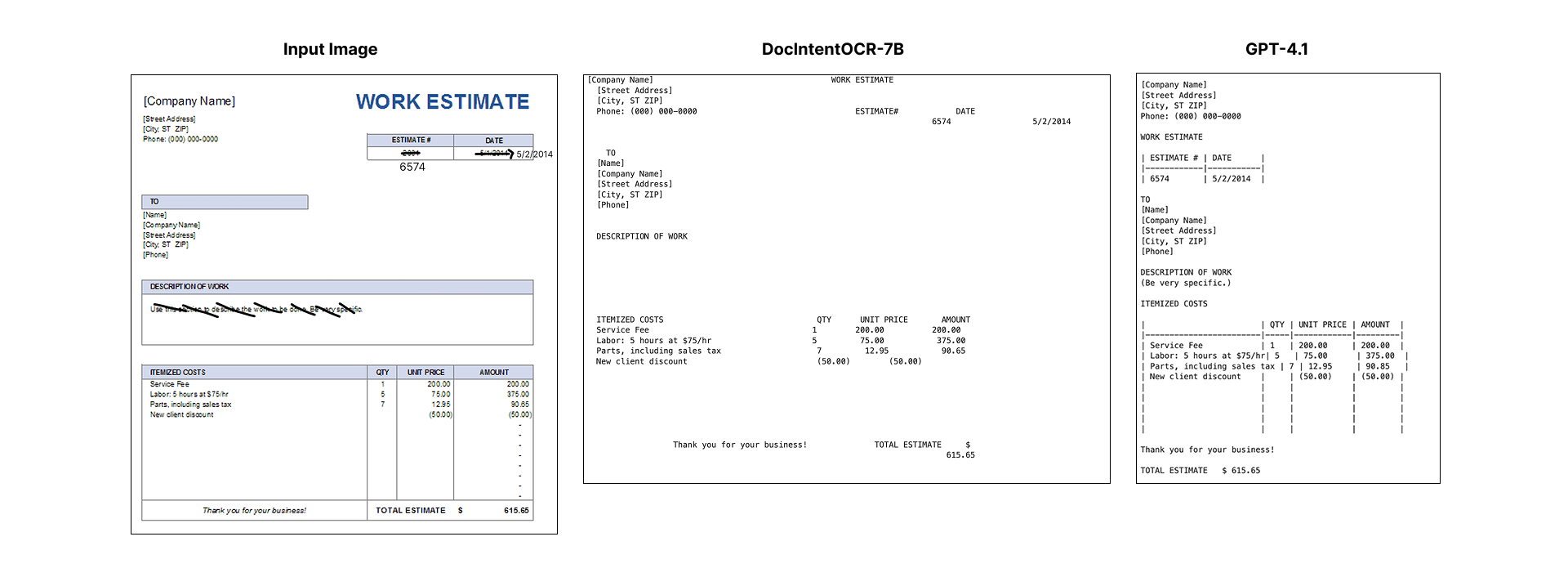}
    \caption{Comparison of outputs from DocIntentOCR-7B and GPT-4.1 on annotated financial documents.}
    \label{fig:intent_ocr}
\end{figure*}

\end{document}